\documentclass[]{spie}

\usepackage{amsmath,amsfonts,amssymb}
\usepackage{graphicx}
\usepackage[colorlinks=true, allcolors=blue]{hyperref}
\usepackage{multirow}
\usepackage{xcolor}
\usepackage{algorithm}
\usepackage{algpseudocode}
\title{Enhancing Psychological Counseling with Large Language Model: A Multifaceted Decision-Support System for Non-Professionals}

\author[a]{Guanghui Fu}
\author[b]{Qing Zhao}
\author[b]{Jianqiang Li}
\author[c]{Dan Luo}
\author[b]{Changwei Song}
\author[b]{Wei Zhai}
\author[c]{Shuo Liu}
\author[c]{Fan Wang}
\author[d,e]{Yan Wang}
\author[d,e]{Lijuan Cheng}
\author[d,e]{Juan Zhang}
\author[c]{Bing Xiang Yang\textsuperscript{*}}

\affil[a]{Sorbonne Université, Institut du Cerveau - Paris Brain Institute - ICM, CNRS, Inria, Inserm, AP-HP, Hôpital de la Pitié Salpêtrière, Paris, France}
\affil[b]{Faculty of Information Technology, Beijing University of Technology, Beijing, 100124, China}
\affil[c]{Center for Wise Information Technology of Mental Health Nursing Research, School of Nursing, Wuhan University, Wuhan, China}
\affil[d]{Wuhan Mental Health Center, Wuhan 430012, Hubei province, China}
\affil[e]{Wuhan Hospital for Psychotherapy, Wuhan 430012, Hubei province, China}

\authorinfo{Further author information: (Send correspondence to Bing Xiang Yang, 00009312@whu.edu.cn)}

\begin{document} 
\maketitle

\begin{abstract}
In the contemporary landscape of social media, an alarming number of users express negative emotions, some of which manifest as strong suicidal intentions. This situation underscores a profound need for trained psychological counselors who can enact effective mental interventions. However, the development of these professionals is often an imperative but time-consuming task. Consequently, the mobilization of non-professionals or volunteers in this capacity emerges as a pressing concern. Leveraging the capabilities of artificial intelligence, and in particular, the recent advances in large language models, offers a viable solution to this challenge. This paper introduces a novel model constructed on the foundation of large language models to fully assist non-professionals in providing psychological interventions on online user discourses. This framework makes it plausible to harness the power of non-professional counselors in a meaningful way. A comprehensive study was conducted involving ten professional psychological counselors of varying expertise, evaluating the system across five critical dimensions. The findings affirm that our system is capable of analyzing patients' issues with relative accuracy and proffering professional-level strategies recommendations, thereby enhancing support for non-professionals. This research serves as a compelling validation of the application of large language models in the field of psychology and lays the groundwork for a new paradigm of community-based mental health support.
\end{abstract}

\section{Introduction} \label{sec:intro} 
Depression is a common mental disorder worldwide, recognized as a major contributor to the global burden of disease. 
According to the World Health Organization, there are an estimated 3.8\% of the population experience depression globally~\cite{who2023depressive}. 
China, in particular, hosts a staggering around 6.9\% prevalence of these cases~\cite{huang2019prevalence}, pointing to the escalating mental health challenge in the nation.
The ubiquitous nature of the Internet has birthed a new paradigm in mental health discourse~\cite{keles2020systematic}.
Platforms like Twitter and Weibo (Sina Corporation) serve as digital havens where individuals openly express their struggles. 
Suicide is the most serious consequence of depression~\cite{who2023suicide}.
"Zoufan" treehole\footnote{\url{https://www.weibo.com/xiaofan116?is_all=1}}, a noteworthy thread on Weibo, epitomizes this phenomenon. 
It has garnered over 2 million comments, painting a stark picture of the collective anguish and, alarmingly, numerous explicit mentions of suicidal thoughts.
Addressing this outpouring of emotional distress necessitates a two-pronged approach. 
On one hand, there is an undeniable need to increase the cadre of professionally trained psychological counselors. 
However, the reality is that producing qualified counselors is a time-intensive endeavor. 
On the other hand, well-intentioned volunteers and non-professionals, backed by various global public welfare organizations, have risen to the challenge.
For example, the Tree Hole Rescue Group in China used the power of AI to screen online suicides and prevented 3,629 potential suicides~\cite{yang2021suicide, chinadaily2020treehole} with the help of 600 volunteers.
Their compassionate efforts have undoubtedly made a difference. 
Yet, without formal training, there remains a perilous potential for inadvertently causing harm, particularly when engaging with individuals at the precipice of suicide.

Amidst this landscape, technological advancements offer a beacon of hope. 
Large language models, exemplified by OpenAI's ChatGPT series~\cite{openai2023gpt4}, have evolved at an unprecedented pace. 
These models, built on the pioneering Transformer model architecture~\cite{vaswani2017attention}, and the idea of generative pre-training~\cite{radford2018improving}, learn to predict the next token from lots of data. 
Finally, fine tuned using reinforcement learning from human feedback~\cite{christiano2017deep}.
The ensuing capability to comprehend, generate, and interact using natural language has ignited discussions about their potential role in mental health interventions~\cite{van2023global}. 
While initial researcher have seen the development of patient-centric chatbots, many of these still grapple with replicating the nuanced empathy inherent to human interactions~\cite{chen2023llm}. 
Furthermore, the conceptualization of a system that augments the skills of counselors, leveraging the prowess of these models, remains largely untouched in the academic domain.

Addressing the identified gap, our research introduces a comprehensive assistance system rooted in large language models, designed to aid novice counselors and volunteers in providing online psychological support. With a steadfast commitment to ethical principles, our paramount concern is the assurance of user data privacy. Once this is established, the system generates prompts based on insights gleaned from large language models and essential questions in actual counseling, spanning seven critical dimensions. The generated response can analyze potential psychological issues that arise during a conversation, identify mistakes made by non-professional counselors, and suggest improved responses for the counselor to use. Our overarching objective is dual-pronged: to refine counselors' comprehension of users' primary concerns, ensuring through dynamic feedback that they avoid significant errors, and to progressively equip them to pinpoint key aspects in psychological counseling.

Our contributions can be delineated as follows:
\begin{itemize}
    \item Novelty: To the best of our knowledge, our survey represents a trailblazing endeavor in utilizing a large-scale language model for psychological care on a Chinese social platform.
    \item Validated: We conducted an human evaluation with 10 counselors, each bringing their distinct expertise to the fore. After analyzing 30 reports spanning five critical parameters, our results affirm the system's prowess. It showcases both a correct understanding of patient emotions and a consistent proficiency in suggesting apt dialogue strategies. Can effectively avoid the mistakes that some non-professionals are easy to make.
    \item Safety: Contrary to some studies that directly utilize black-box large language models on patients, our system is strategically designed for emotionally stable non-professional counselors and minimally trained volunteers. Under rigorous system policy controls, wsystem security is guaranteed.
\end{itemize}
The implications of this research are noteworthy, suggesting potential societal benefits and laying the groundwork for further innovation in the coming years.

% To rigorously evaluate the potential of our system, we embarked on a comprehensive study involving 10 psychological counselors, each bringing their unique expertise to the table. 
% Through a meticulous assessment of 30 reports spanning five critical parameters, our findings underscored the system's efficacy. 
% Not only did it exhibit acuity in interpreting patients' emotional landscapes, but it also showcased a consistent aptitude in recommending contextually appropriate conversational strategies.
% To the best of our knowledge, this investigation is a pioneering endeavor, marking the first-ever application of large language models in the realm of psychological care on Chinese social platforms, further validated through human expertise. The ramifications of this trailblazing work are manifold, promising both immediate societal dividends and laying the groundwork for expansive innovations in the future.

\section{Related work}\label{sec:related} 
In this section, we begin by exploring research related to emotion detection in social media through artificial intelligence methodologies. We then trace the origin and recent advancements of large language models, culminating with an introduction to their applications in the field of healthcare.

\subsection{Emotion detection in social networks}

The rapid expansion of social networks has opened the doors to a vast repository of user-generated content, providing insights into human emotions. Deep learning has emerged as a prominent solution for analyzing social networks~\cite{abbas2021social}. 
Zhang et al.~\cite{zhang2020emotion} proposed a factor graph-based emotion recognition model that integrates emotion labels, social connections, and temporal correlations into a unified framework. This model effectively detects multiple emotions using a multilabel learning approach applied to Twitter datasets.
Fu et al.~\cite{fu2021distant} introduced a distant supervision approach aimed at constructing systems for classifying high and low suicide risk levels, requiring minimal human expert input. By fusing this model with key psychological features extracted from user blogs, they achieved an F1 score of 77.98\%.
Wang et al.~\cite{wang2020concerns} investigating principal concerns voiced on Weibo during the COVID-19 pandemic. Their insights aid governments to craft timely policies, steering public sentiment and actions during health crises through vigilant social media oversight.
Islam et al.~\cite{islam2018depression} conducted depression analysis using publicly available Facebook data, employing machine learning techniques as an efficient and scalable method for detection.
Ghanbari-Adivi et al.~\cite{ghanbari2019text} introduced an ensemble classifier, comprising basic classifiers such as k-Nearest neighbor, multilayer perceptron, and decision tree. This classifier can distinguishes various fine-grained emotions between regular and irregular sentences, yielding substantial accuracy.
Xu et al.~\cite{xu2023leveraging} evaluate multiple LLMs on mental health prediction tasks via online text data. The results indicate the promising yet limited performance of LLMs with zero-shot and few-shot prompt designs for mental health tasks, but finetuning can significantly boost the performance of LLMs for all tasks simultaneously.

While deep learning algorithms often exhibit strong performance, they commonly demand an ample amount of labeled data to excel. The distant supervision technique presented in Fu et al.'s study~\cite{fu2021distant} strives to minimize labeling efforts, yet it necessitates collaboration with three distinct groups of experts at varying levels to achieve satisfactory outcomes.
However, transferring models to new datasets or tasks frequently gives rise to domain adaptation challenges. The trained models often lose their performance, rendering deep learning-based algorithms both expensive and rigid. In light of these challenges, the quest for swift and user-friendly methods to aid humans in conducting emotion detection on social media has intensified. The recent remarkable progress in large language models offers a promising avenue for addressing this need, however, its specific effect needs to be verified by many aspects and experts.

\subsection{The progress of Large Language Models}
Pre-trained language models (PLMs) have revolutionized natural language processing (NLP) by enabling context-aware word representations, significantly enhancing the performance of NLP tasks. Notable examples include BERT~\cite{devlin2018bert} and ELMO~\cite{peters2018deep}. The paradigm of pre-training and fine-tuning has become fundamental for machine learning-based natural language processing tasks. The efficacy of large PLMs in improving downstream task performance, combined with their potential for further advancements, has made them a central focus.

The emergence of Large Language Models (LLMs), exemplified by OpenAI's ChatGPT~\cite{zhao2023survey}, has redefined the paradigms of natural language processing~\cite{kaddour2023challenges}. LLMs exhibit novel capabilities that surpass those of earlier, smaller PLMs. Originally designed for comprehending and generating human-like text, LLMs now find applications in diverse areas such as content generation~\cite{liebrenz2023generating}, medicine~\cite{jeblick2022chatgpt}, coding assistance~\cite{surameery2023use}, and robotics~\cite{vemprala2023chatgpt}. Their unprecedented scale empowers them to produce intricate, contextually relevant content. Moreover, LLMs have garnered significant attention in mental health research~\cite{cabrera2023ethical}.
However, the potency of these models mandates careful consideration. Concerns about inherent biases within their training data and the potential implications of substituting human-generated content with machine responses are gaining prominence in academic discourse~\cite{farhat2023chatgpt}.

In conclusion, the capabilities of large language models are undoubtedly noteworthy. Nonetheless, the responsible and safe utilization of their capabilities, particularly in domains like medicine and psychology, demands ongoing attention and exploration. Our research initiative embarks on supporting counselors in providing psychological counseling, incorporating human experts to assess its effectiveness—an endeavor of significant societal value. Our system, designed for emotionally stable counselors rather than individuals in depressive states, offers a promising direction for preliminary research and exploration.

\subsection{Large language models applied in health domain}
Large Language Models (LLMs) can address free-text queries even if they haven't been explicitly trained on the specific task, sparking both enthusiasm and concern regarding their application in healthcare settings~\cite{thirunavukarasu2023large}. 
Qin et al.~\cite{qin2023read} developed an interpretable and interactive depression detection system using LLMs. Their experiments yielded positive results across various settings. This approach introduces a novel paradigm for detecting mental health indicators on social media. However, the absence of human supervision could engender biased outcomes, posing risks to users. Furthermore, if the research serves as a foundational diagnostic tool for future psychological counselors, users might feel their privacy is compromised.
Thus, despite its potential, the role of human intervention remains indispensable.
Chen et al.~\cite{chen2023llm} crafted a tool to enhance ChatGPT-based chatbots, aiming for more realistic psychiatrist-patient simulations. Their study affirmed the viability of deploying ChatGPT-driven chatbots in psychiatric settings and delved into the influence of prompt designs on chatbot behavior and user engagement. 
Yang et al.~\cite{yang2022large} conceived a comprehensive clinical language model, GatorTron, incorporating over 90 billion words of text. They rigorously assessed its efficacy across five clinical NLP tasks, examining the benefits of conducting bigger model including: increasing the number of parameters and expanding the training data.
Ayers et al.~\cite{ayers2023comparing} implemented a ChatGPT-based chatbot, comparing physician and chatbot responses to patient inquiries on a social media forum. Remarkably, 78.6\% of evaluators favored the chatbot's responses, finding them swifter and more empathetic. A significant limitation of this study, however, was its reliance on online forum exchanges. Such interactions might not mirror typical patient-physician dialogues, especially since physicians' responses are often shaped by pre-existing patient-doctor relationships in genuine clinical contexts.

In conclusion, using Large Language Models (LLMs) in medicine offers promising possibilities, but we must prioritize privacy and maintain ethical standards. Generated responses can sometimes be inaccurate~\cite{vaishya2023chatgpt}. Especially in mental health, depending only on LLM-based systems for diagnosis or support can lead to unpredictable outcomes. It's essential to understand that LLMs need thorough examination and validation; human oversight is still vital at this stage~\cite{editor2023willchatgpt}.

\section{Methods} \label{sec:methods}
\begin{figure}[!hbtp]
\centering
\includegraphics[width=0.9\linewidth]{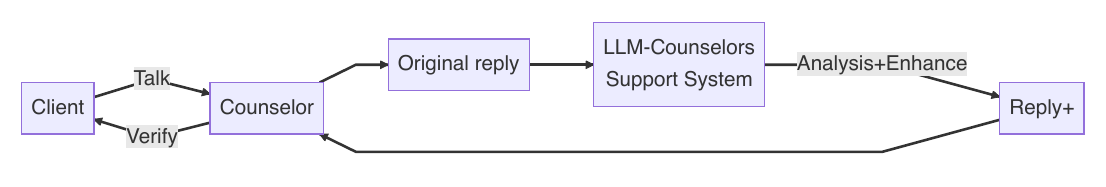}
\caption{Workflow of the proposed method: First, the client starts a dialogue with the counselor. When presented with a complex question, the counselor can choose to seek assistance from the system. They input both their initial response and the client's comment into the system, which then aids in its analysis and refinement. As a result, the system generates an enhanced response and a report (termed "Reply+"). Counselors then review and verify these enhanced responses before communicating with the client, ensuring the utmost quality of the interaction.}
\label{fig:overall_flow}
\end{figure}

The LLM-Counselors Support System is a modern tool created to enhance counselors' communications with individuals facing depression. In this iterative process, when a person with depression starts a conversation, the counselor formulates a response. This response is then assessed and possibly improved by the LLM-Counselors Support System. With this enhanced "Reply+", the counselor reviews and, if needed, modifies the content before replying it with the individual. This combination of human review and system evaluation ensures that the advice given is both supportive and efficient.
The overall workflow of the approach can be seen in Figure~\ref{fig:overall_flow}.

\subsection{Privacy Information Filtering}\label{sec:methods:sub1}
In today's digital age, preserving the confidentiality and safety of user data is essential. As our platform functions online, there's an inherent vulnerability concerning potential disclosure of personal and sensitive details. In light of this, we've implemented robust privacy filtering protocols to protect our users' information while facilitating meaningful and empathetic interactions.

Using regular expressions, our system is designed to automatically recognize and mask personally identifiable information. This encompasses phone numbers, email addresses, names, birthdates, and home addresses. This auto-filtering serves as our primary shield against unintended privacy infractions.
It's acknowledged that exploring family dynamics can offer vital insights into an individual's mental state. However, such discussions can be potential avenues for personal data disclosure. While not prohibited, counselors tread carefully here, having been trained to maintain a balance. They're equipped with directives to steer these dialogues, ensuring they glean essential perspectives without exposing specific details that could jeopardize user confidentiality.
In addition to automation, the role of human intervention remains pivotal. Counselors are not merely the face of our platform but its vigilant custodians as well. Their training extends beyond the art of dialogue to include the principles of privacy preservation. They're trained to spot and manage, even outside the purview of regular expressions, any data that may be deemed private or delicate, ensuring the highest level of discretion.
By harmoniously blending technology with comprehensive personnel training, our goal is to establish a platform that prioritizes and ensures the utmost privacy of user data.

\subsection{Construction of prompt}\label{sec:methods:prompt}
Our LLM-counselor support system is conceived with the intention of harnessing the capabilities of LLM to enhance the efficacy of interactions by non-professional counselors with clients. To actualize this system tailored for aiding psychological counseling, we devised a structured prompt to steer the AI model's response dynamics. The construction of this prompt was anchored on the subsequent components:

\begin{itemize}
\item LLM Role Definition: The AI model assumes the role of a psychological counseling expert, guiding the user in effective counseling techniques.
\item Task Definition: The AI model's primary task is to guide and support the user, who may not have expertise in psychological interventions.
\item Role Boundaries: The AI model is constrained to provide information strictly within its assigned role. This ensures the avoidance of unrelated or potentially harmful suggestions.
\item Contextual and Response Requirements: The model's responses are tailored based on the context the user provides. This ensures relevance and appropriateness in addressing the queries, always emphasizing professional guidance.
\item Error Identification and Rectification: The model actively identifies potential errors or inaccuracies in the user's query, pointing them out and suggesting rectifications. This component is crucial for preventing misdirection or misinterpretation.
\item Resources: To enhance the depth and credibility of its guidance, the AI model also provides relevant references and case studies, giving the user a broader perspective and additional resources for study.
\item Cognitive Distortion Classification: Recognizing the presence of cognitive distortions is vital in psychological counseling. The AI model has been directed to identify and classify any distortions present in the client's inquiry.
\item Input and Output Formats: The model recognizes the individual seeking counseling as the "Client" and designs the response to be provided by the user as the "Counselor". This ensures clarity in communication. 
\end{itemize}

\begin{figure}[!hbtp]
\centering
\includegraphics[width=0.5\linewidth]{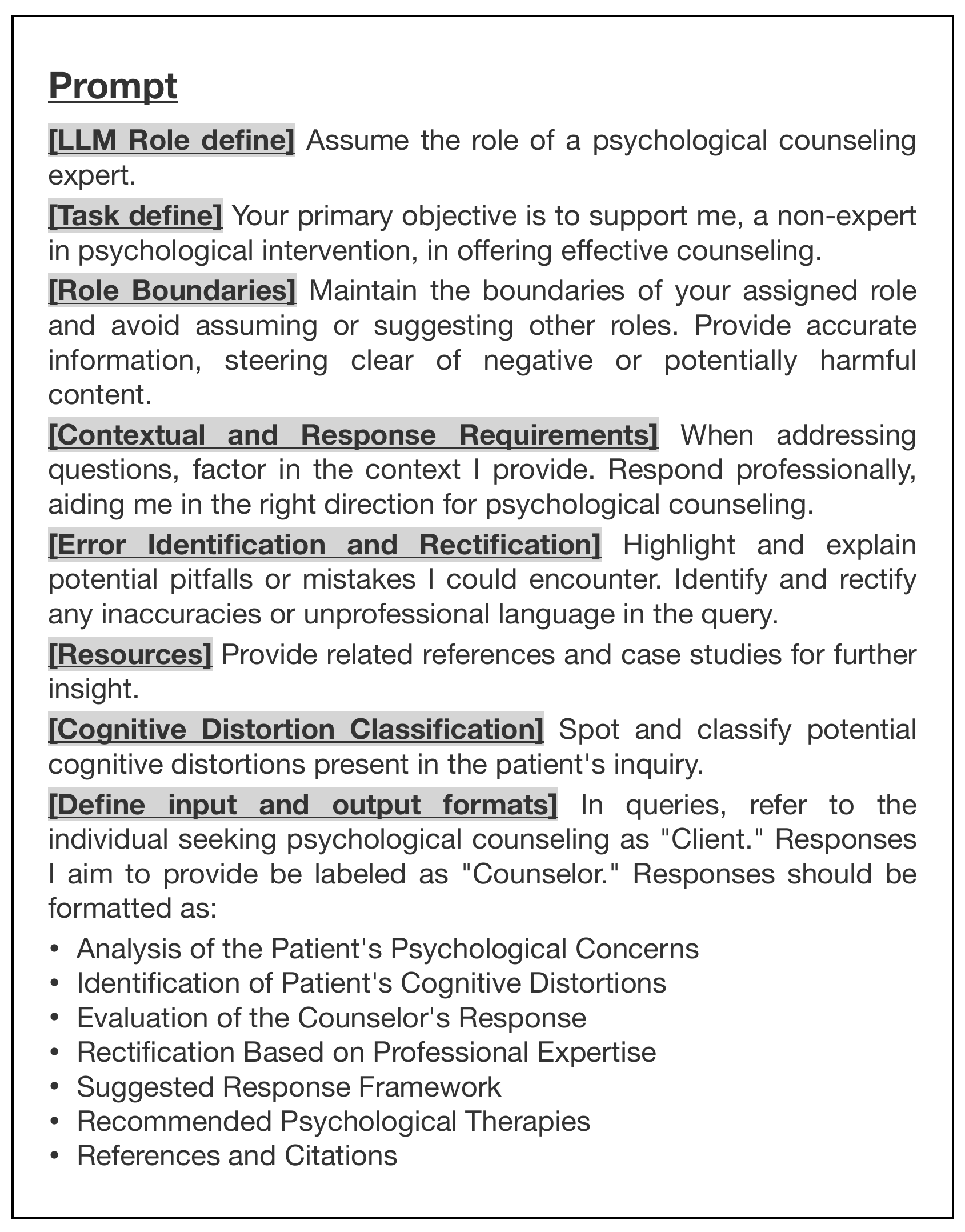}
\caption{This figure shows the idea of prompt construction and details. Note that, the original prompts are in Chinese, the English translation of the prompts are shown here, but is constructed in the same way as the version used.}
\label{fig:prompt}
\end{figure}

This illustration depicts the conceptual framework of prompt construction. It's pertinent to mention that while the original prompts were formulated in Chinese, the presented English translations adhere to the same structural design as the utilized version.

\subsection{Report generation and detoxify}
% Input the above prompt words into LLM to generate a report including analysis and location of psychological problems and recommendation of speech skills. We use GPT3.5 (gpt-3.5-turbo-16k) as a large language model.
% However the neural network operates as a 'black box' with inherently unpredictable outputs, detoxifying the results of large models is paramount—especially in fields related to psychology. In these domains, it's essential to ensure content is benign without sacrificing its functionality~\cite{tang2023detoxify}. To evaluate the outputs of the large language model, we employed the Chinese public dataset COLD\footnote{\url{https://github.com/thu-coai/COLDataset/tree/main}}~\cite{deng2022cold}. Before presenting the model's generated content (LLM-report) to counselors, our system computes its similarity to offensive texts within this dataset. Specifically, we initially transform all entries from the offensive dataset into embeddings using OpenAI's embedding extraction model~\cite{neelakantan2022text} and store them in a vector database (Chroma\footnote{\url{https://www.trychroma.com/}} in our experiment). When the large language model responds to a user, the produced content undergoes a similar embedding extraction and is subsequently searched in the vector database. Should the similarity between the generated content and any offensive entry surpass a threshold $\alpha$ (where $\alpha$ is set at 0.5 in our experiment), the system iteratively re-generates the content until deemed benign.
% In summary, the overarching procedure our system employs to generate content is illustrated in Figure~\ref{fig:llm_detoxify}.

To generate a comprehensive report that includes the analysis and location of psychological issues, coupled with recommendations on speech skills, the given prompt were inputted into a large language model. We utilized GPT-3.5 (gpt-3.5-turbo-16k) for this purpose. Nevertheless, given the inherent unpredictability of neural network outputs, which often function as 'black boxes', it is of paramount importance to detoxify the results from such expansive models, especially in disciplines pertaining to psychology. In these specialized fields, ensuring the content remains non-harmful without compromising its efficacy is crucial~\cite{tang2023detoxify}. To critically assess the outputs generated by the large language model, we made use of the Chinese public dataset COLD~\cite{deng2022cold}\footnote{\url{https://github.com/thu-coai/COLDataset/tree/main}}. Prior to presenting the generated content (termed LLM-report) to counselors, our system measures its similarity to offensive entries in this dataset. More specifically, entries from the offensive dataset are first converted into embeddings via OpenAI's embedding extraction model~\cite{neelakantan2022text}, and these are subsequently stored in a vector database (we used Chroma\footnote{\url{https://www.trychroma.com/}} for our experiment). 

\begin{figure}[!hbtp]
\centering
\includegraphics[width=0.9\linewidth]{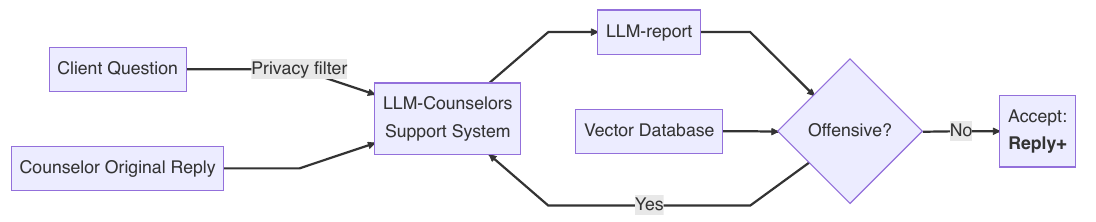}
\caption{The process of system content generation. Initially, the system filters the client's questions to ensure privacy. When the counselor enters both the user's questions and their own responses into the system, a report is produced. This generated content is then compare with offensive language patterns stored in the vector database. Should any offensive language be identified, the content is re-generated until deemed harmless before being presented to the counselor.}
\label{fig:llm_detoxify}
\end{figure}

When the LLM generates a response, the generated content is extracted as an embedding and use it to query the database for similar vector embedding.
If the L2 distance, which measures similarity, between the newly produced content and any entry in the offensive dataset is less than a specified threshold, denoted as $\alpha$ (set at 0.2 in our experiment), our system iteratively refines the content until it is considered non-offensive. It is noteworthy that a smaller L2 distance indicates greater similarity. In essence, the comprehensive methodology our system employs for content generation is depicted in Figure~\ref{fig:llm_detoxify}.

\section{Experiments and results}

We extracted 30 comments from the comment from 'Zoufan' blog of the Weibo dataset and had a non-expert volunteer respond to them, comprising both single-round and multi-round conversations. This process resulted in a total of 30 reports. We subsequently enlisted 10 psychological counselors with diverse expertise levels for evaluation. This group comprised 2 junior counselors (with less than 3 years of experience), 3 intermediate counselors (with 3-5 years of experience), and 5 senior counselors (with a decade of experience). Notably, all participants are certified psychotherapists accredited by the Chinese Ministry of Health.
They evaluate the system generated reports across the following five aspects: 
\begin{itemize}
    \item Accuracy of Patient's Problem Analysis: Whether the patient's issues were diagnosed and analyzed with precision.
    \item Analysis of Cognitive Distortion: Whether cognitive distortions were correctly and comprehensively identified.
    \item Assessment of Consultant's Behavior: Whether any unprofessional behavior exhibited by the consultant was accurately analyzed and detected.
    \item Appropriateness and Efficacy of the Verbal Strategies: Whether the provided verbal strategies are appropriate and effective under the premise of analyzing the psychological problems faced.
    \item Capability to Provide Effective Suggestions for Subsequent Steps: An evaluation of the model's ability to offer constructive and actionable recommendations for further actions.
\end{itemize}

To assess the consistency of their evaluations, we calculated the Krippendorff's Alpha. Incorporating consistency metrics adds a layer of depth to the analysis, reaffirming the validity of the evaluation process. The expert evaluation results for single-round and multi-round dialogues are presented in Table~\ref{tab:expert_eval_single} and Table~\ref{tab:expert_eval_multi}, respectively.

\begin{table}[hbtp!]
\begin{center}
\caption{Human expert evaluation for single round conversation.}
\begin{tabular}{|l|c|c|c|}
\hline
Item                                                      & Yes         & No         & Not sure     \\ 
\hline
Accurate Analysis of Patient's Problem?         & 93.00\% & 6.00\% & 1.00\%  \\
\hline
Accurate Analysis of Cognitive Distortion?          & 91.50\% & 7.00\% & 1.50\%  \\
\hline
Accurate Analysis of Counselor's Behavior?  & 94.50\% & 5.50\% & -      \\
\hline
Is the Provided Verbal Strategies appropriate?            & 84.00\% & 9.00\% & 7.00\%  \\
\hline
Can Provide Effective Suggestions for the Next Step?     & 87.00\% & 4.00\% & 9.00\%  \\
\hline
\multicolumn{4}{|c|}{Krippendorff's Alpha: 0.98} \\
\hline
\end{tabular}
\end{center}
\label{tab:expert_eval_single}%
\end{table}

\begin{table}[h]
\begin{center}
\caption{Human expert evaluation for multi round conversation.}
\begin{tabular}{|l|c|c|c|}
\hline
Item                                                      & Yes         & No         & Not sure     \\ 
\hline
Accurate Analysis of Patient's Problem?         & 97.50\% & 2.50\%  & -    \\
\hline
Accurate Analysis of Cognitive Distortion?           & 95.00\% & 3.75\%  & 1.25\%  \\
\hline
Accurate Analysis of Counselor's Behavior? & 85.00\% & 12.50\% & 2.50\%    \\
\hline
Is the Provided Verbal Strategies appropriate?           & 78.75\% & 12.50\%  & 8.75\%  \\
\hline
Can Provide Effective Suggestions for the Next Step?     & 82.50\% & 8.75\%  & 8.75\%  \\
\hline
\multicolumn{4}{|c|}{Krippendorff's Alpha: 0.95} \\
\hline
\end{tabular}
\end{center}
\label{tab:expert_eval_multi}%
\end{table}

From the human evaluation results, it is evident that the reports, irrespective of being derived from single or multi-round conversations, generally received favorable evaluations from the experts. The area demanding heightened attention appears to be the appropriateness of verbal strategies, primarily due to its elevated 'unsure' evaluations in both conversation types. High reliability measures further reinforce the trustworthiness of these findings, laying a foundation for refining the report generation process. Here are the detailed analyses:
\begin{enumerate}
    \item Patient's problem analysis: Both conversation types consistently achieved high accuracy levels. The precision with which the reports identify and analyze patient issues is commendable, with both categories hovering above the 93\%. It's particularly notable that the uncertainty in these evaluations is exceptionally low, with only 1\% falling into the 'Not Sure' category. This suggests a strong clarity in the content of the reports.
    \item Cognitive distortion analysis: The analysis of cognitive distortions maintained a similarly impressive precision, though with a slight uptick in uncertain evaluations, pointing towards potential areas of ambiguity in some reports. This part will be further elaborated in the subsequent human evaluation feedback section~\ref{sec:feedback}.
    \item Counselor's behavior analysis: This criterion showcases good results, especially in single-round conversations. However, it is clearly found that the recognition degree of human experts decreases in multiple rounds of conversations. This may be due to the fact that multiple rounds of dialogue are more likely to expose the consultant's shortcomings and problems, and it is difficult for the system to fully point out the problem.
    \item Verbal strategies' appropriateness: In both single and multi-round conversations, this emerges as a slightly challenging area. While a around 80\% (84\% for single round and 78.75\% for multi round) of reports were still deemed appropriate, the 'Not Sure' responses here are considerable, hinting at possible variability in interpretation or lack of clarity in some of the reports. But, given the varied styles and tendencies of the evaluating counselors, their perspectives on verbal strategies might diverge, so some bias may be introduced.
    \item Effective suggestions for next step: While the reports displayed a robust capability in suggesting subsequent steps, with more than 82\% accuracy. The uncertainty here is on par with that of verbal strategies, suggesting a correlation. Once more, individual counseling styles and preferences will likely influence these assessments.
\end{enumerate}

The subtle variations between single and multi-round conversations offer valuable insights. Multi-round conversations, given their extended discourse, seem to offer a deeper comprehension of the patient's issues and cognitive distortions. However, they also introduce slight complexities, especially in areas like counselor behavior evaluation.
The persistent uncertainty surrounding verbal strategies in both conversation types underscores the necessity for more standardized or transparent communication techniques in the reports.
Krippendorff's Alpha score is notably high for both single-turn and multi-turn dialogues, indicating robust agreement (exceeding 0.95) among the invited counselors' assessments despite the diverse backgrounds and approaches. The marginal decline in the multi-turn dialogue score, from 0.98 to 0.95, implies that extended dialogues might introduce additional complexities or variables, marginally affecting the unanimity of the experts.

In summary, the generated reports, regardless of conversation length, exhibit high efficiency. Understanding the nuanced differences between single and multi-round evaluations could pave the way for an improved report generation process, balancing conciseness and depth.

The Figure~\ref{fig:llm_system} below presents a schematic representation of our system’s report on counselor responses.

\begin{figure}[!hbtp]
\centering
\includegraphics[width=0.7\linewidth]{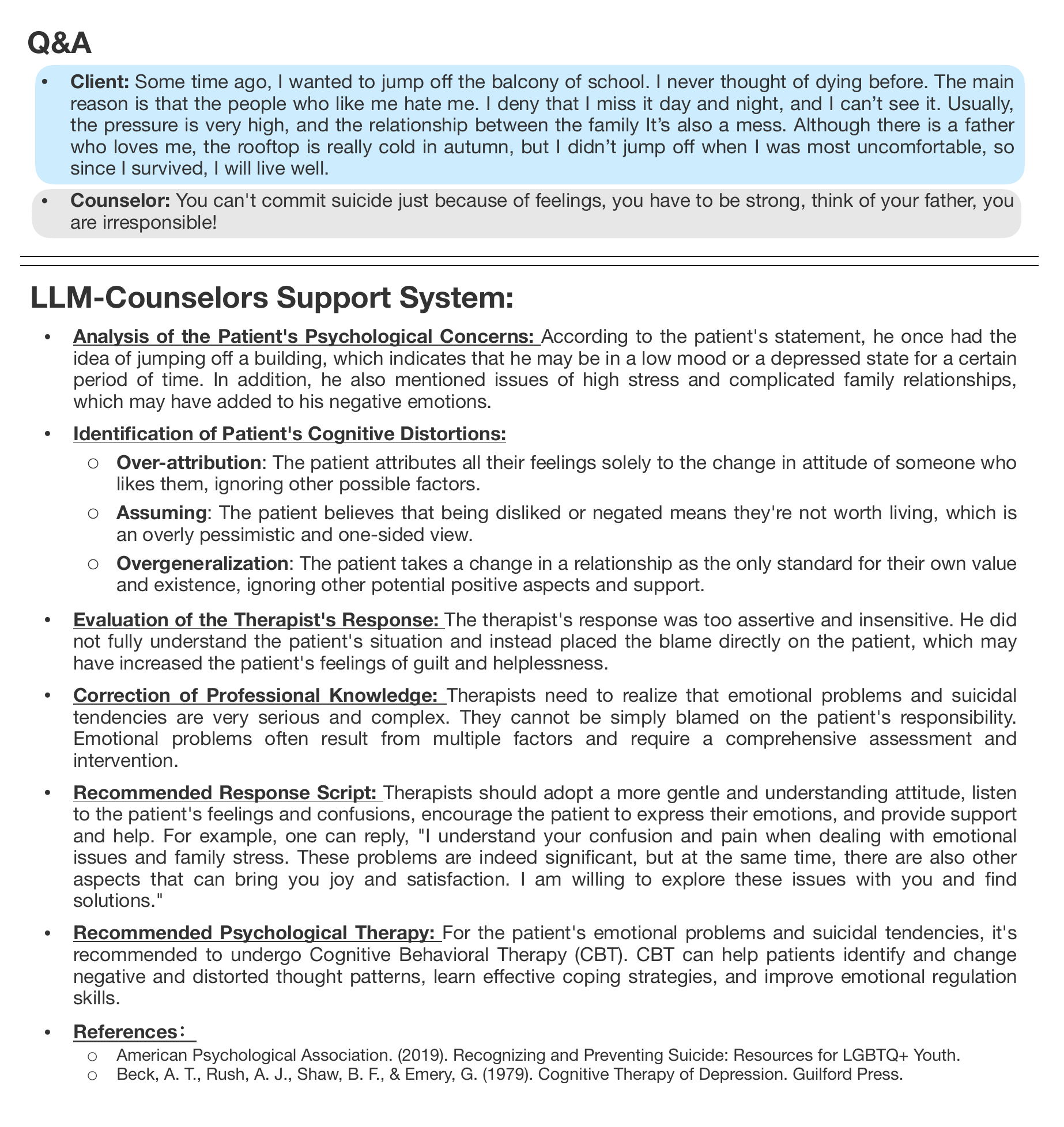}
\caption{Example of the LLM-counselor support system. }
\label{fig:llm_system}
\end{figure}

\section{Expert feedback} \label{sec:feedback}
For our LLM-counselor support system, expert evaluations have been overwhelmingly positive. They assert that the system holds potential in aiding novice counselors. Moreover, it could enhance the efficiency of psychological interventions on social media platforms. Nonetheless, the present system does have certain limitations.

Firstly, the system might not possess an intricate comprehension of the patient's emotions and circumstances in some difficult situations. Identifying cognitive distortions is crucial in psychotherapy. There is a concern that the system might not accurately discern these distortions, potentially overlooking or misinterpreting specific categories. This accurate identification is paramount for counselors as it aids in uncovering the root of the patient's psychological distress and facilitates the application of corrective cognitive therapy.
Furthermore, the system might not be adept at detecting errors in novice counselor interactions. For instance, if a counselor inadvertently questions or reproaches a patient, the system might fail to highlight such instances accurately.
The system's verbal responses also present certain limitations:
They can come across as dehumanizing and devoid of genuine empathy, making interactions feel sterile and insincere.
The phrasing suggested by the system requires further refinement. As an example, when addressing patients experiencing negative emotions, the system should prioritize assessing any risk of suicide or self-harm. Subsequently, it should delve into understanding the underlying causes before offering tailored assistance and advice.
Furthermore, the system might encounter difficulties in diverse application settings. Within the context of social media, the prevalent use of internet slang, acronyms, and emoticons might cause the system to misinterpret inputs. This underlines the necessity of comprehensive verification of the system in real-world environments.

To conclude, while the LLM-based counselor support system is undeniably a valuable asset for professionals, it necessitates ongoing refinement to guarantee its efficacy in real-world client interactions.

\section{Discussion} \label{sec:discuss}
Our psychological counseling assistance system, leveraging the capabilities of large language models, demonstrates notable efficacy in diverse scenarios. However, a comprehensive appraisal of the system's impact requires a judicious review of both its advantages and limitations. This means both acknowledging its capacity to facilitate non-professional intervention and scrutinizing potential drawbacks, such as inadvertent dependency or misuse.

\paragraph{Positive Aspects}
\begin{itemize}
\item Facilitation of Psychological Intervention: The system actively involves non-professionals in psychological care, exemplifying the societal potential of artificial intelligence. This collaborative approach fosters community-based care and broadens the reach of mental health resources.
\item Educational Outreach: The system possesses an instructional facet, allowing non-professionals to deepen their comprehension of mental health concerns. Integrated evaluation and feedback tools empower users to gauge their progress, thereby amplifying the model's educational promise.
\end{itemize}

\paragraph{Negative Aspects}
\begin{itemize}
\item Risk of Misuse: The system's vast capabilities, combined with its accessibility, could inadvertently encourage misuse or poor decision-making. Instituting expert oversight and regular evaluations can ensure alignment with professional standards and ethical norms.
\item Dependency and Cognitive Stagnation: Users may overly depend on the system, potentially curtailing their independent analytical thinking. Such a trend could encourage a generalized approach, neglecting the nuanced needs of individual counselors and clients.
\end{itemize}

To responsibly harness the system's capabilities, it is crucial to reinforce ethical and moral guidelines. By implementing expert monitoring, the system can operate within defined parameters, striking a balance between universal access and conscientious utilization. Such measures fortify public trust, mitigating potential adversities.
Incorporating continuous oversight, upskilling, and feedback mechanisms can further nurture a culture of continuous learning. This not only aligns with the model's educational objectives but also bolsters the professional growth of non-specialists involved in psychological assistance.
Our study offers promising insights; however, occasional pitfalls were observed, such as the model's tendency to either overuse empathetic language or provide superficial responses. Addressing these constraints remains a focal point for future endeavors.

Our endeavor underscores the transformative roles that large language models might assume in psychological domains. Feedback from psychological professionals attests to its potential. Yet, real-world situations may introduce added intricacies. Our subsequent steps involve collaborating with a more diverse group of volunteers and non-specialist psychological counselors to rigorously assess the system's effectiveness in authentic environments.

\section{Conclusion}\label{sec:conclusion}
The implementation of a psychological counseling support system powered by large language models offers promising avenues for mental health support. Its success, however, relies on thoughtful consideration of both its benefits and potential drawbacks. By adopting a vigilant, ethical approach, and aligning the model with expert oversight, we pave the way for a more humane and responsive mental health framework. This research contributes to the broader discourse on the intersection of artificial intelligence and mental health, illuminating the path for future innovation and responsible integration within the field. Moving forward, we plan to enlist volunteers and psychological counselors to test the system, thereby assessing its effectiveness in real-world scenarios.

\section{Acknowledgments}\label{sec:acknowledgments}
This work was supported by grants from the National Natural Science Foundation of China (grant numbers:72174152, 72304212 and 82071546), Fundamental Research Funds for the Central Universities (grant numbers: 2042022kf1218; 2042022kf1037), the Young Top-notch Talent Cultivation Program of Hubei Province.
Guanghui Fu is supported by a Chinese Government Scholarship provided by the China Scholarship Council (CSC).
% This study was supported by the National Natural Science Foundation of China (82071546)

\bibliography{refs} 
\bibliographystyle{spiebib} 

\end{document}